%% file: arxiv.tex
\documentclass{article}

\usepackage[preprint]{corl_2026}

\usepackage{booktabs}
\usepackage{graphicx}
\usepackage{multirow}
\usepackage{natbib}
\usepackage{enumitem}
\usepackage{amsmath}
\usepackage{xcolor}
\usepackage{pgfplots}
\pgfplotsset{compat=1.18}
\usepackage{subcaption}
\usepackage{placeins}
\usepackage{float}
\usepackage{amssymb}
\usepackage{needspace}

\usepackage{titlesec}
\titlespacing*{\section}{0pt}{8pt plus 2pt minus 2pt}{4pt plus 2pt minus 2pt}
\titlespacing*{\subsection}{0pt}{6pt plus 2pt minus 1pt}{3pt plus 1pt minus 1pt}
\titlespacing*{\subsubsection}{0pt}{4pt plus 1pt minus 1pt}{2pt plus 1pt minus 1pt}

\newcommand{\projecturl}{\href{https://yy-gx.github.io/EGR/}{Project website}}

\title{Sensing Which Modality Matters: Evidence-Gated Regularization for Robust VLA Policies}

\author{
  \bfseries Yue Yang$^{1, 2}$ \quad Diego Romeres$^{2}$ \quad Chiori Hori$^{2}$ \quad Gedas Bertasius$^{1}$ \\[1pt]
  \bfseries Daniel Szafir$^{1}$ \quad Siddarth Jain$^{2}$ \\[3pt]
  \mdseries\small $^{1}$Department of Computer Science, University of North Carolina at Chapel Hill \\[0pt]
  \mdseries\small $^{2}$Mitsubishi Electric Research Laboratories (MERL) \\[2pt]
  \mdseries\fontsize{8}{9}\selectfont \texttt{\{yygx,\,gedas,\,daniel.szafir\}@cs.unc.edu} \quad \texttt{diego.romeres@gmail.com} \quad \texttt{\{chori,\,sjain\}@merl.com} \\[2pt]
  \mdseries\small \projecturl
}

\begin{document}
\maketitle
\vspace{-0.18in}   

\begin{abstract}
Vision-Language-Action (VLA) policies fuse multimodal sensory inputs, but training on limited and homogeneous robot demonstrations encourages spurious inter-sensor correlations rather than task-relevant signal, a failure we term modality entanglement. Under real-world occlusions and distractors, this manifests as nuisance sensitivity to corruption of uninformative sensors and single-modality insufficiency when only one informative sensor remains intact.
We propose \textbf{Evidence-Gated Regularization (EGR)}, a modality-agnostic training objective that introduces zero inference-time overhead. EGR derives a per-frame and per-sensor task-relevance signal to gate two state-conditional consistency objectives: invariance on low-evidence sensors, and single-sensor sufficiency on high-evidence ones. We introduce a benchmark based on BEHAVIOR-1K, comprising a fast inference-only diagnostic suite and 47 rollout-based skills targeting modality entanglement. We validate EGR on this benchmark and on two real-robot setups with fundamentally different embodiments: a bi-manual setup with two Kinova arms and three RGB cameras, and a single-arm MELFA ASSISTA setup combining vision and GelSight tactile sensors. EGR improves simulation success rates (SR) from \textbf{12.5\% to 16.4\%} under full modalities (+31\%), from \textbf{9.4\% to 16.5\%} under uninformative-sensor corruption (+75\%), and from \textbf{2.8\% to 6.1\%} under single-sensor fallback (+120\%). Under physical-object distractors, EGR boosts SR from \textbf{30\% to 85\%} on the bi-manual setup (+183\%) and from \textbf{55\% to 70\%} on the tactile setup (+27\%).
\end{abstract}

\input{sections/introduction}
\input{sections/related_work}
\input{sections/methodology}
\input{sections/experimental_results}
\input{sections/conclusion_and_limitation}

\bibliography{main}

\clearpage
\appendix

This appendix supplements the main paper with implementation details, benchmark construction details, statistical analyses, and a theoretical derivation. Section~\ref{app:details} reports hyperparameters. Section~\ref{app:benchmark} reports filter statistics, threshold values, and the sim state replay procedure. Section~\ref{app:stats} formally defines the Suite~1 metric $\Delta$ and reports standard errors, confidence intervals, and significance tests for all numbers in the main paper. Section~\ref{app:theory} proves that the importance-sampled estimators used during training are unbiased.

\section{Implementation Details}
\label{app:details}
\label{app:hyperparams}
The values in this section specify the configuration used to train EGR on BEHAVIOR-1K in simulation and the synthetic corruption operator used to construct the Suite 1 and Suite 2 evaluations. The configuration is identical across all EGR simulation runs reported in the main text.

Table~\ref{tab:hparam-egr} lists the hyperparameters introduced by EGR itself: the two loss weights, the hinge thresholds, the evidence-preprocessing constants, and a flag governing how the per-camera mask channel is treated under corruption.

\begin{table}[H]
\centering
\begin{tabular}{lll}
\toprule
Symbol & Value & Description \\
\midrule
$\lambda_{\mathrm{inv}}$ & 0.2 & weight on $\mathcal{L}_{\mathrm{inv}}$ in the total loss \\
$\lambda_{\mathrm{suff}}$ & 0.1 & weight on $\mathcal{L}_{\mathrm{suff}}$ (denoted \texttt{lambda\_solo} in the released code) \\
$\tau_{\mathrm{low}}$ & 0.2 & hinge knee below which $w_{\mathrm{inv}}$ activates \\
$\tau_{\mathrm{high}}$ & 0.5 & hinge knee above which $w_{\mathrm{suff}}$ activates \\
$\alpha$ & 0.3 & weight of the gated interaction-object contribution in the per-camera score $s_{t,m}$ \\
$\epsilon_{\mathrm{focal}}$ & $10^{-3}$ & interaction-gate threshold on focal-object area $a^{\mathrm{focal}}_{t,m}$ \\
$\epsilon_{\mathrm{global}}$ & $10^{-3}$ & global-gate threshold on $\max_m s_{t,m}$ \\
$\varepsilon$ & $10^{-8}$ & numerical stabilizer in cross-camera normalization \\
\texttt{set\_mask} & False & corrupted cameras retain \texttt{image\_mask}=True so patch tokens stay valid \\
\bottomrule
\end{tabular}
\caption{EGR-specific hyperparameters.}
\label{tab:hparam-egr}
\end{table}

Two distinct corruption operators are used. The training-time operator $\mathcal{C}_m$ injects random rectangular erasures during the computation of $\mathcal{L}_{\mathrm{inv}}$ and $\mathcal{L}_{\mathrm{suff}}$ (Table~\ref{tab:corrupt-train}). The deployment-time operator used in simulation evaluation applies a center-biased black erasure that approximates a large occlusion (Table~\ref{tab:corrupt-eval}). Both operators sample rectangles independently per camera.

\begin{table}[H]
\centering
\begin{tabular}{ll}
\toprule
Parameter & Setting \\
\midrule
Number of rectangles & uniform integer in $\{2,3,4,5\}$ \\
Total erased-area fraction & $\mathrm{Uniform}(0.30,\ 0.80)$ \\
Per-rectangle aspect ratio & $\exp(\mathrm{Uniform}(-1,\ 1)) \approx [0.37,\ 2.72]$ \\
Rectangle placement & top-left position uniform over image \\
Fill value & per-pixel $\mathrm{Uniform}(-1,\ 1)$ noise (images normalized to $[-1,1]$) \\
Per-camera independence & yes \\
\bottomrule
\end{tabular}
\caption{Training-time corruption operator $\mathcal{C}_m$ used inside $\mathcal{L}_{\mathrm{inv}}$ and $\mathcal{L}_{\mathrm{suff}}$.}
\label{tab:corrupt-train}
\end{table}

\begin{table}[H]
\centering
\begin{tabular}{ll}
\toprule
Parameter & Setting \\
\midrule
Number of rectangles & uniform integer in $\{1,\ 2,\ 3\}$ \\
Total erased-area fraction & $\mathrm{Uniform}(0.30,\ 0.70)$ \\
Per-rectangle aspect ratio & $\mathrm{Uniform}(0.5,\ 2.0)$ \\
Rectangle placement & Gaussian about image center with $\sigma = 0.10\,W$ and $0.10\,H$, clipped to fit \\
Fill value & 0 (black) \\
Per-camera independence & yes \\
\bottomrule
\end{tabular}
\caption{Sim-deployment corruption operator used in Suite 1 and Suite 2 to instantiate the NoUseless, UsefulOnly, and SingleUseful regimes.}
\label{tab:corrupt-eval}
\end{table}

Table~\ref{tab:training-config} lists the training-loop configuration. All EGR simulation runs warm-start from the $\pi_{0.5}$ checkpoint provided by~\cite{bai2025openpicometcompetitionsolution}, pre-trained on the 12 BEHAVIOR-1K Challenge tasks; only LoRA adapter weights are updated.

\begin{table}[H]
\centering
\begin{tabular}{ll}
\toprule
Setting & Value \\
\midrule
Batch size & 288 \\
Optimizer & AdamW, $\beta_1=0.9$, $\beta_2=0.95$, $\epsilon=10^{-8}$ \\
Peak learning rate & $2.5 \times 10^{-5}$ \\
LR schedule & cosine decay, 1{,}000-step warmup, decay over 50{,}000 steps, end LR 0 \\
Total training steps & 50{,}000 \\
Weight decay & $10^{-10}$ \\
Gradient clipping & global-norm clip at 1.0 \\
EMA & disabled \\
Numerical precision & bfloat16 model; float32 loss and parameter accumulators \\
Hardware & $2 \times$ NVIDIA B200, FSDP across both devices \\
Dataloader workers & 8 \\
PaliGemma backbone & \texttt{gemma\_2b\_lora} (LoRA on the 2B base) \\
Action expert & \texttt{gemma\_300m\_lora} (LoRA on the 300M base) \\
Action horizon & 32 \\
Trainable parameters & LoRA adapters only; base weights frozen \\
\bottomrule
\end{tabular}
\caption{Training-loop configuration for EGR simulation runs.}
\label{tab:training-config}
\end{table}

\section{Benchmark Construction}
\label{app:benchmark}

\subsection{Filter Pipeline Statistics}
\label{app:filter}
The benchmark curation pipeline starts from the 11 BEHAVIOR-1K Challenge tasks that yield usable annotations. Task 34 is excluded upstream because its annotation is truncated, removing 4 candidate segments before the evidence pipeline is run. The 11 retained tasks contribute 118 candidate segments, 39 navigation (NAV) and 79 manipulation (MAN). Three vision-evidence filters reduce this pool to the 47 segments evaluated in the main text. Filter 1 retains segments whose per-frame global gate $G_t = 1$ on at least 80\% of frames, removing segments in which no camera observes the focal object on a typical frame. Filter 2 retains segments whose head camera is labeled useful, restricting evaluation to scenarios where the head view is informative. Filter 3 enforces a clean wrist-camera contrast: MAN segments are kept only when exactly one wrist is useless (so that the contrast between the useful wrist and the useless wrist is unambiguous), and NAV segments are kept when at least one wrist is useless (a looser criterion, since wrists are typically uninformative during locomotion). Table~\ref{tab:filter-funnel} reports the segment count after each filter, broken out by skill type.

\begin{table}[H]
\centering
\begin{tabular}{lrrr}
\toprule
Stage & All & NAV & MAN \\
\midrule
Input (11 evaluation tasks) & 118 & 39 & 79 \\
After Filter 1 (segment-level global-gate pass) & 103 & 25 & 78 \\
After Filter 2 (head-camera useful) & 99 & 25 & 74 \\
After Filter 3 (wrist-camera ablation criterion) & 47 & 11 & 36 \\
\bottomrule
\end{tabular}
\caption{Segment counts through the three vision-evidence filters.}
\label{tab:filter-funnel}
\end{table}

\subsection{Threshold Values}
\label{app:thresholds}
Two thresholds operate at the per-frame level and drive the per-camera useful/useless classification: $\epsilon_{\mathrm{global}}$ defines when a frame carries evidence at all, and $\mathtt{useless\_threshold}$ defines when a frame counts as not seeing the focal object on a given camera. The remaining two thresholds aggregate frame-level outcomes to segment-level decisions: $\mathtt{g\_threshold}$ on the fraction of frames passing the global gate, and $\mathtt{frac\_threshold}$ on the fraction of frames below $\mathtt{useless\_threshold}$. A camera $m$ is labeled useless on a segment if at least 75\% of its frames satisfy $E_{t,m} < 0.01$, and useful otherwise. Table~\ref{tab:filter-thresholds} lists each threshold, its numeric value, and which filter consumes it.

\begin{table}[H]
\centering
\begin{tabular}{llp{6.5cm}l}
\toprule
Threshold & Value & What it gates & Consumed by \\
\midrule
$\epsilon_{\mathrm{global}}$ & $10^{-3}$ & per-frame global gate, $G_t = 1$ iff $\max_m s_{t,m} > 10^{-3}$ & Filter 1 (defines $G_t$) \\
$\mathtt{g\_threshold}$ & 0.80 & minimum fraction of frames with $G_t = 1$ for a segment to pass & Filter 1 \\
$\mathtt{useless\_threshold}$ & 0.01 & per-frame, per-camera cutoff on $E_{t,m}$ below which the frame counts as not seeing the focal object & Filter 2 and Filter 3 \\
$\mathtt{frac\_threshold}$ & 0.75 & fraction of frames below $\mathtt{useless\_threshold}$ required to label a camera useless on the segment & Filter 2 and Filter 3 \\
\bottomrule
\end{tabular}
\caption{Threshold values used in the filter pipeline.}
\label{tab:filter-thresholds}
\end{table}

\subsection{Sim State Replay Procedure}
\label{app:replay}
The official BEHAVIOR-1K HDF5 stores per-frame state as a slot-keyed serialized tensor; loading it into a Suite 2 evaluation scene with a different object set mis-aligns the slots, and a single-frame jump-in additionally loses hidden physics state such as joint velocities, contact constraints, and particle-system bookkeeping. We therefore replay each task's first episode from frame 0 in \texttt{VISUAL\_ONLY} mode and, at each segment start frame, call \texttt{og.sim.dump\_state(serialized=False)} to save a name-keyed snapshot covering all object poses, robot and gripper joint configurations, and particle-system state. At evaluation time, Suite 2 loads the snapshot directly into the evaluation scene and runs one zero-gravity settling step so PhysX re-establishes contact constraints before the policy begins acting.

\section{Metrics and Statistical Analysis}
\label{app:stats}

\subsection{Definition of $\Delta$ for Suite 1}
\label{app:delta_definition}
For each segment with relevant action dimensions $\mathcal{D}$ and frames $\mathcal{F}$,
\begin{equation}
\Delta = \underbrace{\frac{1}{|\mathcal{F}|}\sum_{t\in\mathcal{F}}\|\pi_\theta(o_t^{\mathrm{corrupt}}) - a_t^\star\|_{\mathcal{D}}^2}_{\mathrm{MSE}_{\mathrm{corrupt}}} - \underbrace{\frac{1}{|\mathcal{F}|}\sum_{t\in\mathcal{F}}\|\pi_\theta(o_t^{\mathrm{full}}) - a_t^\star\|_{\mathcal{D}}^2}_{\mathrm{MSE}_{\mathrm{full}}},
\label{eq:delta-mse}
\end{equation}
where $a_t^\star$ is the ground-truth action chunk and $\mathcal{D}$ selects the robot base dimensions for NAV segments and the torso plus both arm dimensions for MAN segments.

\subsection{Standard Errors and Confidence Intervals}
\label{app:se-ci}
This subsection reports per-cell uncertainty for every number used in Tables 1 and 2 and in Fig.~4 of the main paper. Aggregated means use $\mathrm{SE} = \mathrm{SD}/\sqrt{n}$ across the natural sampling unit: segments for the simulation suites and tasks for the real-robot platforms. Binomial rates additionally report the Wilson 95\% interval over the pooled trial count. Suite 1 Ratios additionally report a 95\% bootstrap interval from 10{,}000 paired resamples over segments. All computations use a fixed seed.

\paragraph{Suite 1.}
Each cell reports the Full and corrupt MSE means with standard errors and the Ratio with a bootstrap 95\% interval; the EGR row has the lowest Ratio in every panel.
\begin{table}[H]
\centering
\small
\resizebox{\textwidth}{!}{ %
\begin{tabular}{llrccc}
\toprule
Panel & Method & $n$ & Full MSE (mean $\pm$ SE) & Corrupt MSE (mean $\pm$ SE) & Ratio \% (mean $\pm$ SE) [95\% CI] \\
\midrule
NAV / SingleUseful & vanilla & 11 & $0.0616 \pm 0.0105$ & $0.0809 \pm 0.0115$ & $+31.35 \pm 9.33$ $[+16.8, +53.4]$ \\
NAV / SingleUseful & ModDrop & 11 & $0.0853 \pm 0.0130$ & $0.1172 \pm 0.0172$ & $+37.42 \pm 14.38$ $[+15.6, +72.3]$ \\
NAV / SingleUseful & EGR & 11 & $0.0780 \pm 0.0115$ & $0.0903 \pm 0.0128$ & $+15.83 \pm 4.66$ $[+8.0, +26.3]$ \\
MAN / NoUseless & vanilla & 36 & $0.0020 \pm 0.0002$ & $0.0022 \pm 0.0002$ & $+12.74 \pm 4.93$ $[+4.0, +23.3]$ \\
MAN / NoUseless & ModDrop & 36 & $0.0024 \pm 0.0002$ & $0.0026 \pm 0.0002$ & $+10.90 \pm 2.65$ $[+6.0, +16.4]$ \\
MAN / NoUseless & EGR & 36 & $0.0023 \pm 0.0003$ & $0.0024 \pm 0.0002$ & $+5.07 \pm 2.67$ $[+0.4, +10.9]$ \\
MAN / UsefulOnly & vanilla & 36 & $0.0020 \pm 0.0002$ & $0.0052 \pm 0.0007$ & $+158.54 \pm 28.36$ $[+105.0, +216.4]$ \\
MAN / UsefulOnly & ModDrop & 36 & $0.0024 \pm 0.0002$ & $0.0056 \pm 0.0008$ & $+137.73 \pm 19.82$ $[+99.4, +176.7]$ \\
MAN / UsefulOnly & EGR & 36 & $0.0023 \pm 0.0003$ & $0.0051 \pm 0.0007$ & $+121.82 \pm 25.10$ $[+74.0, +172.4]$ \\
\bottomrule
\end{tabular}}
\caption{Suite 1 per-cell statistics. Standard errors are computed across segments; Ratio confidence intervals are from 10{,}000 paired bootstrap resamples over segments.}
\label{tab:se-suite1}
\end{table}

\paragraph{Suite 2 (NAV).}
Per-segment SR is a binomial proportion over $n=20$ trials, and we report the cross-segment mean $\pm$ SE together with a bootstrap 95\% CI over $n=11$ segments.
\begin{table}[H]
\centering
\small
\begin{tabular}{llrcc}
\toprule
Condition & Method & $n_{\text{seg}}$ & SR (mean $\pm$ SE) & 95\% CI \\
\midrule
Full & vanilla & 11 & $42.27 \pm 14.78$ & $[15.9, 70.0]$ \\
Full & ModDrop & 11 & $40.00 \pm 14.13$ & $[15.5, 66.8]$ \\
Full & EGR & 11 & $40.91 \pm 14.47$ & $[15.0, 68.2]$ \\
SingleUseful & vanilla & 11 & $15.45 \pm 9.28$ & $[1.8, 34.5]$ \\
SingleUseful & ModDrop & 11 & $15.45 \pm 9.18$ & $[1.8, 34.5]$ \\
SingleUseful & EGR & 11 & $37.27 \pm 13.76$ & $[13.6, 63.6]$ \\
\bottomrule
\end{tabular}
\caption{Suite 2 NAV success rate per cell. Means and SEs are computed across $n_{\text{seg}}=11$ segments; CIs are from 10{,}000 bootstrap resamples over segments.}
\label{tab:se-suite2-nav}
\end{table}

\paragraph{Suite 2 (MAN).}
For each cell, the mean is computed across $n_{\text{seg}}=36$ MAN segments; each per-segment value summarizes 20 trials. We report SE and bootstrap 95\% CI across segments for all three metrics.
\begin{table}[H]
\centering
\small
\begin{tabular}{llcccccc}
\toprule
 & & \multicolumn{2}{c}{SR (\%)} & \multicolumn{2}{c}{EC (\%)} & \multicolumn{2}{c}{EE distance (m)} \\
\cmidrule(lr){3-4}\cmidrule(lr){5-6}\cmidrule(lr){7-8}
Condition & Method & mean $\pm$ SE & 95\% CI & mean $\pm$ SE & 95\% CI & mean $\pm$ SE & 95\% CI \\
\midrule
Full & vanilla & $12.50 \pm 3.77$ & $[6.0, 20.4]$ & $52.78 \pm 6.15$ & $[40.8, 64.6]$ & $0.249 \pm 0.049$ & $[0.16, 0.35]$ \\
Full & ModDrop & $7.50 \pm 3.17$ & $[2.5, 14.6]$ & $40.97 \pm 6.24$ & $[28.9, 52.9]$ & $0.287 \pm 0.048$ & $[0.20, 0.39]$ \\
Full & EGR & $16.39 \pm 4.10$ & $[8.9, 24.7]$ & $50.00 \pm 5.95$ & $[38.5, 61.4]$ & $0.257 \pm 0.049$ & $[0.17, 0.36]$ \\
NoUseless & vanilla & $9.44 \pm 3.41$ & $[3.8, 16.8]$ & $47.22 \pm 6.01$ & $[35.7, 58.6]$ & $0.264 \pm 0.045$ & $[0.18, 0.36]$ \\
NoUseless & ModDrop & $6.81 \pm 3.05$ & $[1.9, 13.3]$ & $38.89 \pm 5.96$ & $[27.5, 50.4]$ & $0.292 \pm 0.047$ & $[0.21, 0.39]$ \\
NoUseless & EGR & $16.53 \pm 4.21$ & $[9.2, 25.1]$ & $53.89 \pm 6.07$ & $[41.9, 65.6]$ & $0.246 \pm 0.046$ & $[0.16, 0.34]$ \\
UsefulOnly & vanilla & $2.78 \pm 2.78$ & $[0.0, 8.3]$ & $22.78 \pm 6.29$ & $[11.2, 35.1]$ & $0.447 \pm 0.059$ & $[0.34, 0.57]$ \\
UsefulOnly & ModDrop & $2.78 \pm 2.78$ & $[0.0, 8.3]$ & $22.36 \pm 6.34$ & $[10.8, 35.0]$ & $0.470 \pm 0.064$ & $[0.35, 0.60]$ \\
UsefulOnly & EGR & $6.11 \pm 3.05$ & $[1.4, 12.9]$ & $30.14 \pm 6.01$ & $[18.9, 42.2]$ & $0.366 \pm 0.052$ & $[0.27, 0.47]$ \\
\bottomrule
\end{tabular}
\caption{Suite 2 MAN per-cell statistics for the three reported metrics. $n_{\text{seg}}=36$ for every cell.}
\label{tab:se-suite2-man}
\end{table}

\paragraph{Real bi-manual platform.}
Each (task, condition, method) cell contains $n=10$ trials, pooled across 4 tasks to $N=40$ trials per (condition, method). We report the pooled rate with its Wilson 95\% interval over the 40 trials, together with the cross-task SE over $n_{\text{task}}=4$.
\begin{table}[H]
\centering
\small
\begin{tabular}{lcccccc}
\toprule
 & \multicolumn{3}{c}{Baseline} & \multicolumn{3}{c}{EGR} \\
\cmidrule(lr){2-4}\cmidrule(lr){5-7}
Condition & Pooled $n/N$ & Wilson 95\% CI & Cross-task SE & Pooled $n/N$ & Wilson 95\% CI & Cross-task SE \\
\midrule
Clean & 28/40 & $[0.55, 0.82]$ & $0.70 \pm 0.07$ & 32/40 & $[0.65, 0.90]$ & $0.80 \pm 0.07$ \\
NoUseless & 21/40 & $[0.37, 0.67]$ & $0.53 \pm 0.05$ & 36/40 & $[0.77, 0.96]$ & $0.90 \pm 0.07$ \\
UsefulOnly & 10/40 & $[0.14, 0.40]$ & $0.25 \pm 0.10$ & 29/40 & $[0.57, 0.84]$ & $0.73 \pm 0.10$ \\
RealDistractor & 12/40 & $[0.18, 0.45]$ & $0.30 \pm 0.15$ & 34/40 & $[0.71, 0.93]$ & $0.85 \pm 0.09$ \\
\bottomrule
\end{tabular}
\caption{Real bi-manual platform SR per condition. The pooled count $n$ aggregates 4 tasks $\times$ 10 trials; the cross-task SE is over $n_{\text{task}}=4$ task-level SRs.}
\label{tab:se-real-bimanual}
\end{table}

\paragraph{Real tactile platform.}
SR per (condition, method) cell pools the 2 tasks $\times$ 10 trials into $N=20$ trials.
\begin{table}[H]
\centering
\small
\begin{tabular}{lcccc}
\toprule
 & \multicolumn{2}{c}{Baseline} & \multicolumn{2}{c}{EGR} \\
\cmidrule(lr){2-3}\cmidrule(lr){4-5}
Condition & Pooled $n/N$ & Wilson 95\% CI & Pooled $n/N$ & Wilson 95\% CI \\
\midrule
Clean & 18/20 & $[0.70, 0.97]$ & 17/20 & $[0.64, 0.95]$ \\
SingleUseful & 8/20 & $[0.22, 0.61]$ & 18/20 & $[0.70, 0.97]$ \\
RealDistractor & 11/20 & $[0.34, 0.74]$ & 14/20 & $[0.48, 0.85]$ \\
\bottomrule
\end{tabular}
\caption{Real tactile platform SR per condition, pooled across the 2 tasks.}
\label{tab:se-real-tactile-sr}
\end{table}

\subsection{Significance Tests}
\label{app:significance}
For the simulation suites we use the one-sided paired Wilcoxon signed-rank test on per-segment values, with EGR as the test arm and one of the two baselines as the reference. The alternative hypothesis follows the metric's direction: for Suite 1 Ratio we test EGR $\Delta <$ baseline $\Delta$ where $\Delta = \mathrm{MSE}_{\mathrm{corrupt}} - \mathrm{MSE}_{\mathrm{full}}$; for Suite 2 SR and EC we test EGR $>$ baseline; for Suite 2 EE distance we test EGR $<$ baseline. For the real-robot platforms we use Fisher's exact test (two-sided) on the pooled trial counts. Significance markers throughout: $^{*}\, p<0.05$, $^{**}\, p<0.01$, $^{***}\, p<0.001$.

\needspace{7cm}
\paragraph{Suite 1.}
EGR's per-segment $\Delta$ is significantly smaller than vanilla in all three panels and smaller than ModDrop in two of the three; the MAN NoUseless comparison against ModDrop is marginal ($p = 0.061$).
\begin{table}[H]
\centering
\small
\begin{tabular}{lllrr}
\toprule
Panel & Comparison & $n$ & $W$ & $p$ \\
\midrule
NAV / SingleUseful & EGR vs vanilla & 11 & $14.0$ & $0.0261$$^{*}$ \\
NAV / SingleUseful & EGR vs ModDrop & 11 & $10.0$ & $0.0105$$^{*}$ \\
MAN / NoUseless & EGR vs vanilla & 36 & $235.0$ & $0.0400$$^{*}$ \\
MAN / NoUseless & EGR vs ModDrop & 36 & $248.0$ & $0.0606$ \\
MAN / UsefulOnly & EGR vs vanilla & 36 & $213.0$ & $0.0181$$^{*}$ \\
MAN / UsefulOnly & EGR vs ModDrop & 36 & $200.0$ & $0.0107$$^{*}$ \\
\bottomrule
\end{tabular}
\caption{Suite 1 paired Wilcoxon signed-rank tests on per-segment $\Delta$. One-sided, EGR $\Delta <$ baseline $\Delta$.}
\label{tab:sig-suite1}
\end{table}

\paragraph{Suite 2 (NAV).}
With only 11 NAV segments the paired Wilcoxon attains its floor of $p = 2^{-4} = 0.0625$ under SingleUseful; all four segments on which the methods differ favor EGR, so the limitation is statistical power rather than effect direction or consistency.
\begin{table}[H]
\centering
\small
\begin{tabular}{llrrrr}
\toprule
Condition & Comparison & $n_{\text{seg}}$ & nonzero diffs & $W$ & $p$ \\
\midrule
Full & EGR vs vanilla & 11 & 3 & $1.0$ & $0.8750$ \\
Full & EGR vs ModDrop & 11 & 3 & $4.0$ & $0.3750$ \\
SingleUseful & EGR vs vanilla & 11 & 4 & $10.0$ & $0.0625$ \\
SingleUseful & EGR vs ModDrop & 11 & 4 & $10.0$ & $0.0625$ \\
\bottomrule
\end{tabular}
\caption{Suite 2 NAV paired Wilcoxon. With only 11 segments and few nonzero differences, the smallest attainable $p$ is $2^{-(\text{nonzero})}$. Under SingleUseful all 4 nonzero per-segment differences favor EGR, attaining the floor $p = 2^{-4} = 0.0625$ for both baseline comparisons.}
\label{tab:sig-suite2-nav}
\end{table}

\paragraph{Suite 2 (MAN).}
The Full-condition SR gain is significant against ModDrop ($p < 0.001$) and marginal against vanilla ($p = 0.065$); it is a secondary effect, since EGR primarily targets the corruption conditions, where NoUseless and UsefulOnly are significant against both baselines.
\begin{table}[H]
\centering
\small
\begin{tabular}{lllr}
\toprule
Metric & Condition & Comparison & $p$ \\
\midrule
SR & Full & EGR vs vanilla & $0.0649$ \\
SR & Full & EGR vs ModDrop & $0.0003$$^{***}$ \\
SR & NoUseless & EGR vs vanilla & $0.0055$$^{**}$ \\
SR & NoUseless & EGR vs ModDrop & $0.0002$$^{***}$ \\
SR & UsefulOnly & EGR vs vanilla & $0.0033$$^{**}$ \\
SR & UsefulOnly & EGR vs ModDrop & $0.0033$$^{**}$ \\
EC & Full & EGR vs vanilla & $0.8079$ \\
EC & Full & EGR vs ModDrop & $0.0029$$^{**}$ \\
EC & NoUseless & EGR vs vanilla & $0.0186$$^{*}$ \\
EC & NoUseless & EGR vs ModDrop & $0.0001$$^{***}$ \\
EC & UsefulOnly & EGR vs vanilla & $0.0057$$^{**}$ \\
EC & UsefulOnly & EGR vs ModDrop & $0.0073$$^{**}$ \\
EE distance & Full & EGR vs vanilla & $0.7540$ \\
EE distance & Full & EGR vs ModDrop & $0.0051$$^{**}$ \\
EE distance & NoUseless & EGR vs vanilla & $0.1990$ \\
EE distance & NoUseless & EGR vs ModDrop & $0.0025$$^{**}$ \\
EE distance & UsefulOnly & EGR vs vanilla & $0.0000$$^{***}$ \\
EE distance & UsefulOnly & EGR vs ModDrop & $0.0000$$^{***}$ \\
\bottomrule
\end{tabular}
\caption{Suite 2 MAN paired Wilcoxon signed-rank tests across $n_{\text{seg}}=36$ segments. SR/EC use $H_1:$ EGR $>$ baseline; EE distance uses $H_1:$ EGR $<$ baseline.}
\label{tab:sig-suite2-man}
\end{table}

\needspace{6cm}
\paragraph{Real bi-manual.}
All three corruption conditions (NoUseless, UsefulOnly, RealDistractor) are significant at $p < 0.001$ against the baseline, while Clean shows no significant difference, consistent with EGR not targeting clean performance.
\begin{table}[H]
\centering
\small
\begin{tabular}{lcccr}
\toprule
Condition & Baseline $n/N$ & EGR $n/N$ & Odds ratio & $p$ \\
\midrule
Clean & 28/40 & 32/40 & $1.71$ & $0.4391$ \\
NoUseless & 21/40 & 36/40 & $8.14$ & $0.0004$$^{***}$ \\
UsefulOnly & 10/40 & 29/40 & $7.91$ & $0.0000$$^{***}$ \\
RealDistractor & 12/40 & 34/40 & $13.22$ & $0.0000$$^{***}$ \\
\bottomrule
\end{tabular}
\caption{Real bi-manual Fisher's exact test (two-sided) on pooled trial counts across the 4 tasks.}
\label{tab:sig-real-bimanual}
\end{table}

\paragraph{Real tactile.}
On RealDistractor the tactile platform pools only $N = 20$ trials and the EGR advantage (14/20 vs 11/20) does not reach significance ($p = 0.51$); the same RealDistractor condition on the bi-manual platform, with $N = 40$ trials, is strongly significant ($p < 0.001$, Table~\ref{tab:sig-real-bimanual}), indicating that the tactile result is underpowered rather than absent.
\begin{table}[H]
\centering
\small
\begin{tabular}{lcccr}
\toprule
Condition & Baseline $n/N$ & EGR $n/N$ & Odds ratio & $p$ \\
\midrule
Clean & 18/20 & 17/20 & $0.63$ & $1.0000$ \\
SingleUseful & 8/20 & 18/20 & $13.50$ & $0.0022$$^{**}$ \\
RealDistractor & 11/20 & 14/20 & $1.91$ & $0.5145$ \\
\bottomrule
\end{tabular}
\caption{Real tactile platform Fisher's exact test (two-sided) on pooled SR. RealDistractor shows a positive point estimate that does not reach significance at $N=20$ pooled trials.}
\label{tab:sig-real-tactile-sr}
\end{table}

\section{Unbiased Importance-Sampled Estimator for $\mathcal{L}_{\mathrm{inv}}$ and $\mathcal{L}_{\mathrm{suff}}$}
\label{app:theory}
\label{app:importance-sampling}
The per-sample contribution of $\mathcal{L}_{\mathrm{inv}}$ in Eq.~(2) of the main paper is
\begin{equation}
\ell_b^{\mathrm{inv}} = G_t \sum_{m \in \mathcal{M}} w_{\mathrm{inv}}(E_{b,m}) \cdot \big\| v_\theta(o_b) - v_\theta(\tilde{o}_b^{(m)}) \big\|_2^2 .
\end{equation}
Let $W_b^{\mathrm{inv}} = \sum_{m \in \mathcal{M}} w_{\mathrm{inv}}(E_{b,m})$. When $W_b^{\mathrm{inv}} > 0$, define the categorical distribution
\begin{equation}
p_{b,m}^{\mathrm{inv}} = \frac{w_{\mathrm{inv}}(E_{b,m})}{W_b^{\mathrm{inv}}}, \qquad \sum_{m \in \mathcal{M}} p_{b,m}^{\mathrm{inv}} = 1 .
\end{equation}
Rewriting the per-sample loss using this distribution,
\begin{equation}
\ell_b^{\mathrm{inv}} = G_t \cdot W_b^{\mathrm{inv}} \cdot \sum_{m \in \mathcal{M}} p_{b,m}^{\mathrm{inv}} \cdot \big\| v_\theta(o_b) - v_\theta(\tilde{o}_b^{(m)}) \big\|_2^2 = G_t \cdot W_b^{\mathrm{inv}} \cdot \mathbb{E}_{m \sim p_b^{\mathrm{inv}}}\!\left[\big\| v_\theta(o_b) - v_\theta(\tilde{o}_b^{(m)})\big\|_2^2\right] .
\end{equation}
Our estimator draws a single sample $m^\star_b \sim p_b^{\mathrm{inv}}$ and forms
\begin{equation}
\hat{\ell}_b^{\mathrm{inv}} = G_t \cdot W_b^{\mathrm{inv}} \cdot \big\| v_\theta(o_b) - v_\theta(\tilde{o}_b^{(m^\star_b)}) \big\|_2^2 .
\end{equation}
Taking expectation over $m^\star_b$,
\begin{equation}
\mathbb{E}_{m^\star_b \sim p_b^{\mathrm{inv}}}\!\left[\hat{\ell}_b^{\mathrm{inv}}\right] = G_t \cdot W_b^{\mathrm{inv}} \cdot \mathbb{E}_{m \sim p_b^{\mathrm{inv}}}\!\left[\big\| v_\theta(o_b) - v_\theta(\tilde{o}_b^{(m)})\big\|_2^2\right] = \ell_b^{\mathrm{inv}} .
\end{equation}
The $W_b^{\mathrm{inv}}$ multiplier is the importance-sampling correction; without it the estimator would be biased downward by a factor of $W_b^{\mathrm{inv}}$. When $W_b^{\mathrm{inv}} = 0$, no modality on sample $b$ triggers the invariance hinge; we then fall back to a uniform $1/|\mathcal{M}|$ prior over modalities for sampling, and the loss contribution remains zero through the $W_b^{\mathrm{inv}}$ multiplier. Summing $\hat{\ell}_b^{\mathrm{inv}}$ over the batch gives the estimator $\hat{\mathcal{L}}_{\mathrm{inv}}$ of Eq.~(5) of the main paper, so $\mathbb{E}[\hat{\mathcal{L}}_{\mathrm{inv}}] = \mathcal{L}_{\mathrm{inv}}$. The derivation for $\hat{\mathcal{L}}_{\mathrm{suff}}$ is identical, with $w_{\mathrm{suff}}$ and $W_b^{\mathrm{suff}}$ replacing their invariance counterparts.
This estimator preserves the expected gradient of the full-sum loss while reducing the per-step cost from $2|\mathcal{M}| + 1$ forward passes to three total (one clean, one for $\hat{\mathcal{L}}_{\mathrm{inv}}$, one for $\hat{\mathcal{L}}_{\mathrm{suff}}$), independent of the number of modalities.

\end{document}

%% file: sections/introduction.tex
\vspace{-7pt}
\section{Introduction}
\label{sec:intro}
\vspace{-7pt}

Robot learning has made substantial progress, enabling embodied agents to perform an increasingly broad range of complex tasks~\cite{chi2025diffusion, brohan2022rt, yang2024enhancing, sivamayil2023systematic}. Vision-Language-Action (VLA) policies have recently emerged as a leading paradigm for general-purpose robot learning. These policies typically build on large-model backbones and rely on early multimodal fusion to integrate visual, linguistic, and action-relevant information~\cite{sapkota2025vision, kim2024openvla, intelligence2025pi}. However, early fusion can encourage shortcut learning, causing models to exploit spurious cross-modal correlations rather than task-relevant signals~\cite{geirhos2020shortcut, dancette2021beyond}. VLA policies inherit this vulnerability, and the problem is amplified for robot demonstration data. Compared with the web-scale corpora used to train multimodal foundation models, robot demonstrations are typically far smaller and less diverse in operators, environments, and objects~\cite{o2024open, yang2024arcade}. We refer to the resulting failure mode as \emph{modality entanglement}: a phenomenon in which VLA policies develop spurious dependencies on sensory modalities that are not task-relevant at the current state.

\begin{figure}[tb]
  \centering
  \includegraphics[width=\linewidth]{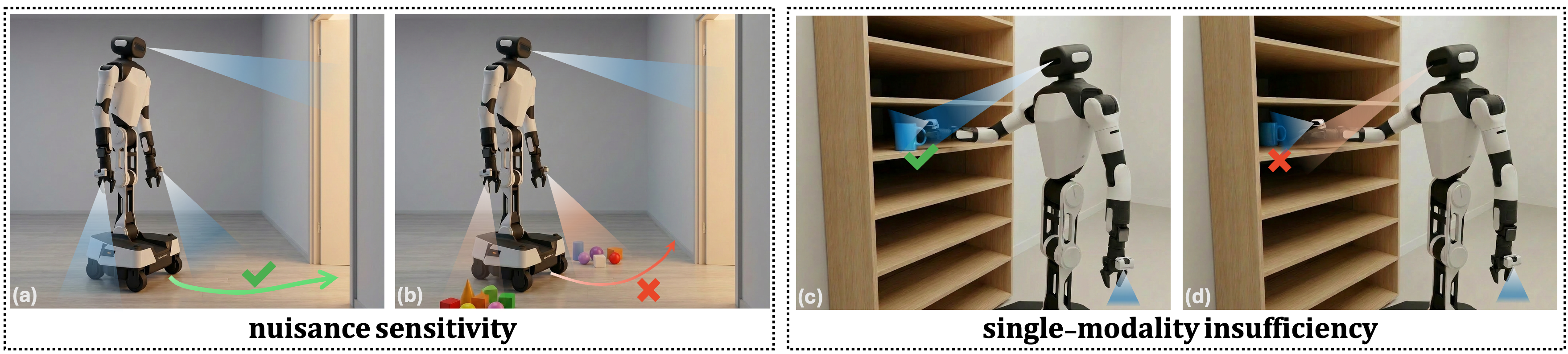}
  
  \caption{Two failure modes of modality entanglement. \textbf{(a-b) Nuisance sensitivity:} distractors in a task-irrelevant wrist camera break navigation despite an unchanged head camera. \textbf{(c-d) Single-modality insufficiency:} when a deeper-placed target makes the shelf board occlude the head camera, the policy stops despite an intact wrist view.}
  \label{fig:teaser}
\end{figure}

We illustrate modality entanglement effect through two representative, state-dependent failure modes (Fig.~\ref{fig:teaser}). As a first example, consider navigation, where the head camera provides the task-relevant information while the two downward-facing wrist cameras observe only the floor (Fig.~\ref{fig:teaser}(a)). At deployment, adding physical distractors to the wrist-camera views alone can stall the policy, even though the head-camera observation remains unchanged (Fig.~\ref{fig:teaser}(b)). This suggests that the policy has spuriously associated the specific floor appearance in the wrist views with being on the correct path. We refer to this behavior as \emph{nuisance sensitivity}: degradation caused by perturbations to sensors that, at the current state, contain no task-relevant signal. As a second example, consider reaching into a shelf, where both the head camera and the right wrist camera initially observe the target (Fig.~\ref{fig:teaser}(c)). At deployment, the target is placed deeper inside the shelf cell, causing the upper board to occlude the head-camera view while the right wrist camera continues to see the target. Nevertheless, the policy stops or behaves erratically (Fig.~\ref{fig:teaser}(d)). This suggests that the policy has not learned to treat the wrist camera alone as sufficient evidence for continuing the task. We refer to this behavior as \emph{single-modality insufficiency}: when multiple sensors carry task-relevant signal but all are corrupted except one, the policy fails to fall back to that remaining sensor.

At its core, modality entanglement reflects a missing capability: the policy should be able to decide, at each frame, which sensors provide task-relevant evidence. Existing approaches differ in the signals they use, or fail to use, to guide this decision. Random modality dropout~\cite{liu2017learning} and adversarial multi-sensor consistency~\cite{guo2025robustness} target generic robustness to sensor availability by treating sensors and timesteps uniformly. As a result, they improve tolerance to missing or perturbed inputs but do not explicitly perform state-dependent modality selection. Attention-based methods allow the policy to learn such selection end-to-end, but they rely on the same limited and homogeneous demonstration data that gives rise to modality entanglement in the first place. Consequently, the learned attention can inherit the same spurious co-occurrences as the underlying policy~\cite{hao2023masked}. TacVLA~\cite{zhang2026tacvla} instead introduces an explicit architectural gate, activating tactile tokens only when contact is detected. However, this gate is specialized to a single modality and cue: tactile sensing and binary contact. Extending this idea to arbitrary sensor types would require redesigning the gating mechanism itself. What remains missing is a modality-selection method that is explicit, grounded in task structure rather than fragile demonstration statistics, and decoupled from any particular sensor modality.

We propose \textbf{Evidence-Gated Regularization (EGR)}, a training objective that satisfies these three desiderata. For each training frame, EGR assigns an evidence score that measures task-grounded signal from each sensor. Rather than learning this score end-to-end from demonstrations, we derive it from properties of the task. The score then gates two complementary consistency objectives: invariance to perturbations of low-evidence sensors, which encourages the policy to ignore uninformative inputs, and single-sensor sufficiency for high-evidence sensors, which encourages the policy to act on an informative signal. In this way, the evidence score serves as a soft inductive bias, allowing the policy to exploit additional useful structure. A key feature of EGR is its separation between modality-specific evidence construction and modality-agnostic regularization. The former defines task-grounded signal per sensor; the latter applies uniform regularization once that signal is computed. This design allows EGR to be applied consistently across heterogeneous sensor modalities while leaving the base VLA architecture unchanged. Because EGR is imposed only during training, it incurs no inference-time overhead. In summary, we contribute the following:

\noindent\textbf{(1)} We formalize modality entanglement in VLA policies and characterize it through two state-dependent failure modes---nuisance sensitivity and single-modality insufficiency. 
\vspace{-4pt}

\noindent\textbf{(2)} We introduce Evidence-Gated Regularization (EGR), a modality-agnostic training objective that leverages per-frame, per-sensor task-relevance evidence to gate two state-conditional consistency objectives, while adding zero inference-time overhead.
\vspace{-4pt}

\noindent\textbf{(3)} We construct a BEHAVIOR-1K benchmark for modality entanglement, with a fast inference-only diagnostic suite and 47 rollout-based skills. On this benchmark, EGR improves success rates from \textbf{12.5\% to 16.4\%} under full modalities (+31\%), from \textbf{9.4\% to 16.5\%} under uninformative-sensor corruption (+75\%), and from \textbf{2.8\% to 6.1\%} under single-sensor fallback (+120\%).
\vspace{-4pt}

\noindent\textbf{(4)} We validate EGR on two real-robot setups with fundamentally different embodiments: a bi-manual platform with two Kinova arms and three RGB cameras, and a single-arm MELFA ASSISTA platform with one RGB camera and two GelSight tactile sensors. Under physical-object distractors, EGR boosts SR from \textbf{30\% to 85\%} on the bi-manual setup (+183\%) and from \textbf{55\% to 70\%} on the tactile setup (+27\%).

%% file: sections/related_work.tex
\vspace{-7pt}
\section{Related Work}
\label{sec:related-work}
\vspace{-7pt}
\subsection{Shortcut Learning in Multimodal LLMs}
\label{subsec:shortcut-learning}
\vspace{-5pt}

Current Vision-Language-Action (VLA) policies and broader multimodal large language models (MLLMs)~\cite{zitkovich2023rt,kim2024openvla,intelligence2025pi,zhang2026tacvla,huang2025tactile} combine large pretrained backbones with multimodal sensory input. Their generalization to unseen conditions depends critically on whether the model learns task-relevant signals or spurious cross-modal correlations. Geirhos et al.~\cite{geirhos2020shortcut} characterized this as shortcut learning, where models rely on decision rules that work on the training distribution but fail under distribution shift. Dancette et al.~\cite{dancette2021beyond} introduced VQA-CE to expose multimodal shortcuts involving question-image co-occurrence. Recent work characterizes similar effects at MLLM scale~\cite{hosseini2025seeing,cai2025diagnosing,wu2025beyond} and in VLA models~\cite{fei2025libero,fang2026vision}. The phenomenon we characterize, where a VLA binds its policy to sensors that are uninformative at the current state, is a distinct form of inter-sensor shortcut that this prior line has not isolated.

\vspace{-5pt}
\subsection{Modality Robustness in Multi-Sensor Policies}
\label{subsec:modality-robustness}
\vspace{-5pt}

Beyond robotics, multimodal learning has long observed that jointly training on multiple modalities can leave some under-utilized or cause over-reliance on the dominant one~\cite{wang2019multimodalhard,peng2022ogm,du2023unimodal}, motivating balancing and missing-modality-robust techniques~\cite{wu2024missingmodalitysurvey,reza2024robust}. In robot manipulation, recent work weights modalities per stage or state via learned attention~\cite{feng2024play}. In VLA models, recent methods improve robustness through run-time distractor inpainting~\cite{hancock2025run}, attention-pathway restructuring~\cite{zhang2026focusvla}, supervised attention specialization~\cite{jia2026guidedvla}, and training on diverse visual clutters~\cite{yang2025boss, yang2026lilo}. These methods rely on architectural changes, run-time intervention, or end-to-end learned attention, none grounding modality selection in per-frame task structure.

%% file: sections/methodology.tex
\vspace{-7pt}
\section{Methodology}
\label{sec:method}
\vspace{-7pt}

EGR decouples per-frame sensor relevance from its effect on the policy. It first estimates how task-relevant each sensor is at each frame through a per-sensor evidence score (Sec.~\ref{subsec:evidence}), then uses those scores uniformly across the regularization objectives that shape policy learning (Sec.~\ref{subsec:egr-loss}). This separation allows the same framework to handle qualitatively different sensors while keeping inference cost identical to that of the base policy.

\vspace{-5pt}
\subsection{Problem Setup}
\label{subsec:problem_setup}
\vspace{-5pt}

A VLA policy $\pi_\theta(a_t\mid o_t)$ maps a multi-sensor observation $o_t = \{o_t^m\}_{m \in \mathcal{M}}$ to an action distribution, where $\mathcal{M}$ is the set of sensor modalities. The policy is trained via imitation learning on demonstration data $\mathcal{D}$. We use $o_t^m$ to denote the observation in which only modality $m$ is retained and all other modalities are corrupted. We say that $\pi_\theta$ exhibits \textbf{modality entanglement} at $(t, m)$ when its behavioral dependence on modality $m$ is misaligned with the modality’s actual importance for task completion. This misalignment can arise in two complementary ways. First, under \textbf{nuisance sensitivity}, modality $m$ at frame $t$ is task-irrelevant, perturbing it does not change the task success rate,
yet perturbing $m$ changes the policy’s action distribution. Second, under \textbf{single-modality insufficiency}, modality $m$ at frame $t$ is sufficient on its own but other task-relevant modalities coexist, executing $\pi_\theta$ using only $m$ preserves task success rate, yet the policy's action distribution differs when only $m$ is preserved compared to when all sensors are available.

\vspace{-5pt}
\subsection{Evidence as a Task-Relevance Indicator}
\label{subsec:evidence}
\vspace{-5pt}

The evidence score measures how much task-relevant signal a given sensor carries at each frame. For cameras, evidence reflects the visibility of task-relevant geometry; for tactile sensors, it reflects task-relevant contact patterns. More generally, evidence may be computed from any task-relevant signal available during training, whether obtained from task structure or learned representations.

Formally, for each frame $t$ and sensor $m \in \mathcal{M}$, we assign an evidence score $E_{t,m} \in [0, 1]$. Scores are normalized across modalities at each frame, so they capture relative rather than absolute importance. We treat $E_{t,m}$ as a heuristic measure of task relevance, rather than ground truth. It indicates which failure mode (Sec.~\ref{subsec:problem_setup}) the current frame is most susceptible to, thereby enabling selective regularization. The following two subsubsections instantiate this score for vision and tactile sensing.

\vspace{-5pt}
\subsubsection{Vision Instantiation}
\vspace{-5pt}

Vision evidence measures the amount of task-relevant geometry visible to each camera. Intuitively, a camera provides stronger evidence when focal or interaction objects occupy a larger visible area in its view. We assume access to per-frame instance segmentation, together with task annotations that identify focal objects, i.e., the objects being manipulated, and interaction objects, i.e., the receptacles or targets with which the focal objects interact as specified by the task. For example, in a “put coke can into trash bin” task, the coke can is the focal object and the trash bin is the interaction object. For camera $m$ at frame $t$, let $A^{\mathrm{focal}}_{t,m,k} \in [0,1]$ denote the pixel-area ratio of focal object $k$ and $A^{\mathrm{inter}}_{t,m,k} \in [0,1]$ the area ratio of interaction object $k$. We aggregate these object-level quantities into per-frame focal and interaction visibility scores, $a^{\mathrm{focal}}_{t,m} = \sum_{k} A^{\mathrm{focal}}_{t,m,k}$ and $a^{\mathrm{inter}}_{t,m} = \sum_{k} A^{\mathrm{inter}}_{t,m,k}$.

Using aggregated area alone can be misleading: large interaction objects, such as tables, may dominate the evidence in cameras that do not observe the focal object, thereby falsely inflating their scores. To prevent this, we include interaction objects only when the focal object is visible in the same camera, using an interaction gate $g_{t,m} = \mathbb{1}[a^{\mathrm{focal}}_{t,m} > \epsilon_{\mathrm{focal}}]$. The per-camera interaction score combines the focal contribution with the gated interaction contribution, $s_{t,m} = a^{\mathrm{focal}}_{t,m} + \alpha \cdot g_{t,m} \cdot a^{\mathrm{inter}}_{t,m}$, where the hyperparameter $\alpha$ controls the influence of interaction objects.

Cross-camera normalization yields the final evidence $E_{t,m} = s_{t,m} / (\max_{m'} s_{t,m'} + \varepsilon)$, and a global gate $G_t = \mathbb{1}[\max_{m} s_{t,m} > \epsilon_{\mathrm{global}}]$ disables both regularization losses on search frames where no camera carries meaningful evidence, leaving the base imitation loss to drive policy behavior.

\vspace{-5pt}
\subsubsection{Tactile Instantiation}
\vspace{-5pt}

Tactile evidence is contact-based, with two regimes defined by the structure of the task. In the sparse-contact regime, which includes skills such as grasping, pressing, and insertion, the occurrence of contact is itself the task-relevant event. Evidence reduces to a binary contact indicator $E_{t,\mathrm{tactile}} = \mathbb{1}[\mathrm{contact\ at\ frame\ } t]$, similar in spirit to the contact-gating in prior work~\cite{zhang2026tacvla}.
In the rich-contact regime, contact persists throughout the sub-task. Binary contact alone provides no frame-level discriminative signal. We therefore introduce a task-feature operator, $\phi_{\mathrm{task}}$ that maps the current tactile observation to a similarity score against the target feature, yielding $E_{t,\mathrm{tactile}} = \phi_{\mathrm{task}}(o^{\mathrm{tactile}}_t) \in [0,1]$. The specific form of $\phi_{\mathrm{task}}$ is task-dependent. In our evaluation, the target feature is a localized tactile pattern, such as a crack on a surface, and $\phi_{\mathrm{task}}$ compares against reference signatures from demonstrations.

\vspace{-5pt}
\subsection{Evidence-Gated Regularization Objectives}
\label{subsec:egr-loss}
\vspace{-5pt}
\begin{figure}[!htbp]
  \centering
  \includegraphics[width=\linewidth]{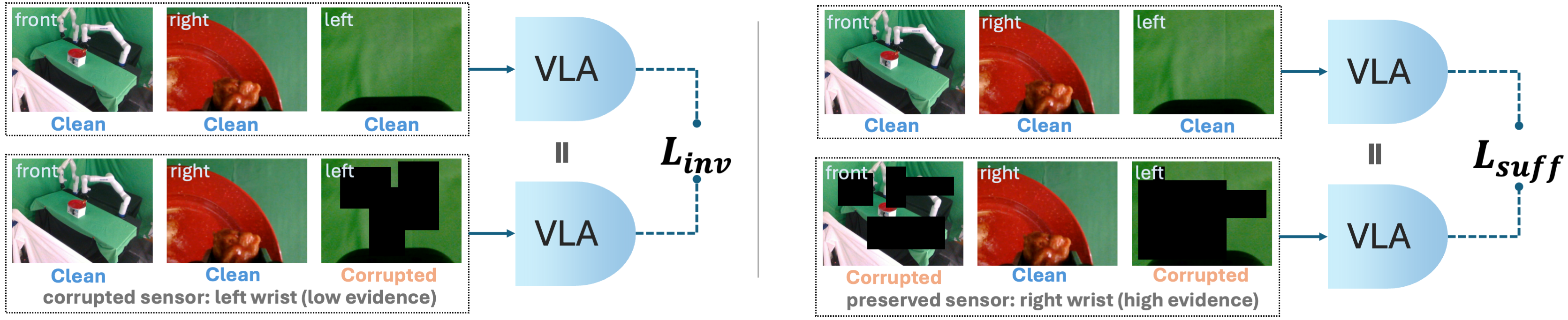}
  \caption{The two evidence-gated loss terms. \textbf{Left:} $\mathcal{L}_{\mathrm{inv}}$ enforces invariance when a low-evidence sensor is corrupted. \textbf{Right:} $\mathcal{L}_{\mathrm{suff}}$ enforces sufficiency when only a high-evidence sensor is preserved.}
  \label{fig:method}
\end{figure}

The evidence score $E_{t,m}$ links the two failure modes formalized in Sec.~\ref{subsec:problem_setup} to the training objective. When evidence is low, it activates an invariance objective on modality $m$, addressing nuisance sensitivity. When evidence is high, it activates a sufficiency objective on $m$, addressing single-modality insufficiency. For intermediate evidence levels, training defaults to the base imitation loss.

We realize this gating through hinge weights anchored at two thresholds $\tau_{\mathrm{low}} < \tau_{\mathrm{high}}$:
\begin{equation}
w_{\mathrm{inv}}(E_{t,m}) = \max\!\left(0,\, \frac{\tau_{\mathrm{low}} - E_{t,m}}{\tau_{\mathrm{low}}}\right), \qquad w_{\mathrm{suff}}(E_{t,m}) = \max\!\left(0,\, \frac{E_{t,m} - \tau_{\mathrm{high}}}{1 - \tau_{\mathrm{high}}}\right).
\label{eq:hinges}
\end{equation}
Between $\tau_{\mathrm{low}}$ and $\tau_{\mathrm{high}}$ lies an ambiguous zone in which neither loss applies. To probe each loss, we introduce a corruption operator $\mathcal{C}_m(\cdot)$ that acts on modality $m$. This yields two corrupted observations: $\tilde{o}^{(m)}_t$, in which modality $m$ alone is corrupted by $\mathcal{C}_m$ and the other modalities are kept intact, and $o^m_t$, defined as in Sec.~\ref{subsec:problem_setup}, in which all modalities other than $m$ are corrupted. We instantiate $\mathcal{C}_m$ using random erasing on all visual inputs, including both camera RGB images and the RGB images produced by GelSight tactile sensors. However, the framework is agnostic to the specific choice of corruption.

The invariance loss targets the nuisance-sensitivity failure mode formalized in Sec.~\ref{subsec:problem_setup}. When modality $m$ provides little evidence, perturbations to $m$ should not alter the policy's action (Fig.~\ref{fig:method}, left). Let $v_\theta$ denote the flow-matching velocity prediction of the underlying $\pi_{0.5}$ policy~\cite{intelligence2025pi}. The global gate restricts the sum to informative frames, and the hinge weight $w_{\mathrm{inv}}$ activates only on low-evidence sensors; the resulting invariance loss is defined in Eq.~\eqref{eq:loss-inv}.
\begin{equation}
\mathcal{L}_{\mathrm{inv}} = \sum_{t,\; m \in \mathcal{M}} G_t \cdot w_{\mathrm{inv}}(E_{t,m}) \cdot \big\| v_\theta(o_t) - v_\theta(\tilde{o}^{(m)}_t) \big\|_2^2.
\label{eq:loss-inv}
\end{equation}

The single-sensor sufficiency loss directly addresses the single-modality insufficiency failure mode formalized in Sec.~\ref{subsec:problem_setup}. When evidence on modality $m$ is high, the policy should produce the same action when only $m$ is preserved (i.e., the observation $o^m_t$ defined in Sec.~\ref{subsec:problem_setup}) as when all sensors are available (Fig.~\ref{fig:method}, right). The hinge weight $w_{\mathrm{suff}}$ activates only on high-evidence sensors; the resulting sufficiency loss is defined in Eq.~\eqref{eq:loss-suff}.
\begin{equation}
\mathcal{L}_{\mathrm{suff}} = \sum_{t,\; m \in \mathcal{M}} G_t \cdot w_{\mathrm{suff}}(E_{t,m}) \cdot \big\| v_\theta(o_t) - v_\theta(o^m_t) \big\|_2^2.
\label{eq:loss-suff}
\end{equation}

The final training objective combines the underlying flow-matching imitation loss $\mathcal{L}_{\mathrm{flow}}$ with the two evidence-gated objectives:
\begin{equation}
\mathcal{L} = \mathcal{L}_{\mathrm{flow}} + \lambda_{\mathrm{inv}} \mathcal{L}_{\mathrm{inv}} + \lambda_{\mathrm{suff}} \mathcal{L}_{\mathrm{suff}}.
\label{eq:total-loss}
\end{equation}
Here $\lambda_{\mathrm{inv}}, \lambda_{\mathrm{suff}} \geq 0$ balance the two regularizers.

\vspace{-5pt}
\subsection{Efficient Training via Importance Sampling}
\label{subsec:efficient-training}
\vspace{-5pt}
Computing $\mathcal{L}_{\mathrm{inv}}$ and $\mathcal{L}_{\mathrm{suff}}$ as written in Eqs.~\eqref{eq:loss-inv} and~\eqref{eq:loss-suff} sums over all $|\mathcal{M}|$ modalities, requiring $2|\mathcal{M}|$ additional corrupted forward passes through $\pi_\theta$ per training step. We reduce this overhead with an unbiased importance-sampled estimator. For each sample $b$ in the batch, let $W_b^{\mathrm{inv}} = \sum_{m \in \mathcal{M}} w_{\mathrm{inv}}(E_{b,m})$ and define the categorical distribution $p_{b,m}^{\mathrm{inv}} = w_{\mathrm{inv}}(E_{b,m}) / W_b^{\mathrm{inv}}$. We sample a single modality $m^\star_b \sim p_b^{\mathrm{inv}}$, evaluate the per-modality loss term only on $m^\star_b$, and rescale by $W_b^{\mathrm{inv}}$, yielding the estimator in Eq.~\eqref{eq:loss-inv-is}.
\begin{equation}
\hat{\mathcal{L}}_{\mathrm{inv}} = \sum_b G_t \cdot W_b^{\mathrm{inv}} \cdot \big\| v_\theta(o_b) - v_\theta(\tilde{o}_b^{(m^\star_b)}) \big\|_2^2.
\label{eq:loss-inv-is}
\end{equation}
The same construct yields $\hat{\mathcal{L}}_{\mathrm{suff}}$ using $w_{\mathrm{suff}}$ and $W_b^{\mathrm{suff}}$. Both estimators are unbiased: $\mathbb{E}[\hat{\mathcal{L}}_{\mathrm{inv}}] = \mathcal{L}_{\mathrm{inv}}$ and $\mathbb{E}[\hat{\mathcal{L}}_{\mathrm{suff}}] = \mathcal{L}_{\mathrm{suff}}$ (see Appendix). Sampling proportional to hinge weights concentrates compute on modalities that actually contribute to the loss for each sample. The training overhead drops from $2|\mathcal{M}|$ to two extra forward passes per step regardless of the number of modalities.

%% file: sections/experimental_results.tex
\FloatBarrier
\section{Experiments and Results}
\label{sec:experiments}
\vspace{-7pt}
Our evaluation is structured around three questions: \textbf{(Q1)} Does modality entanglement, as formalized in Sec.~\ref{subsec:problem_setup}, emerge in trained VLA policies? \textbf{(Q2)} Does EGR mitigate this entanglement? and \textbf{(Q3)} Do the benefits of EGR transfer to real-robot platforms with different sensor configurations?

\vspace{-5pt}
\subsection{Implementation Details}
\label{subsec:impl}
\vspace{-5pt}
We instantiate EGR on $\pi_{0.5}$~\cite{intelligence2025pi}, augmenting its flow-matching imitation loss with $\mathcal{L}_{\mathrm{inv}}$ and $\mathcal{L}_{\mathrm{suff}}$ (Sec.~\ref{sec:method}). Vision evidence is computed from BEHAVIOR-1K's per-frame instance segmentation in simulation, and from Grounding DINO v2~\cite{liu2024grounding} with manually-prompted bounding boxes followed by SAM2~\cite{ravi2025sam} offline for real-robot training data; segmentation is used at training time only. The corruption operator $\mathcal{C}_m$ is instantiated as random erasing on all visual inputs (including GelSight RGB) during training, center erasing for sim deployment as a controlled diagnostic that approximates a large physical occlusion, and physical distractors for real-robot deployment. 

\vspace{-5pt}
\subsection{Benchmark}
\label{subsec:benchmark}
\vspace{-5pt}
Our benchmark provides 47 curated skill segments (11 navigation ``NAV" + 36 manipulation ``MAN") derived from BEHAVIOR-1K, together with two evaluation suites and the metrics for diagnosing modality entanglement at both the action level and the task-completion level.

\vspace{-5pt}
\subsubsection{Evaluation}
\label{subsubsec:evaluation}
\vspace{-5pt}

We evaluate using two suites over the same set of benchmark segments. Suite 1 is an inference-only diagnostic that measures per-frame action deviation under modality perturbations, without rollout-based execution, making it inexpensive to compute. Suite 2 is a rollout-based evaluation that measures task success through real executions, providing a task-completion-level assessment that complements Suite 1’s action-level diagnostics.

We evaluate four corruption regimes across both suites, using consistent terminology throughout the paper. \textbf{Clean} leaves all sensors intact. \textbf{NoUseless} corrupts the useless modality to test nuisance sensitivity (Sec.~\ref{subsec:problem_setup}). \textbf{UsefulOnly} corrupts all modalities except the single useful one to test single-modality insufficiency (Sec.~\ref{subsec:problem_setup}). \textbf{SingleUseful}, used only in setups with exactly one useful sensor, corrupts all remaining sensors and thus combines NoUseless and UsefulOnly into a single regime. For setups with multiple useful sensors, we evaluate Clean, NoUseless, and UsefulOnly separately to isolate the two failure modes. For setups with exactly one useful sensor, we evaluate Clean and SingleUseful. In Suite 2, each segment is rolled out 20 times to estimate task success rate.

\vspace{-5pt}
\subsubsection{Baselines and Metrics}
\label{subsubsec:baselines-metrics}
\vspace{-5pt}

\paragraph{Baselines.} We compare EGR against two baselines. The first is vanilla $\pi_{0.5}$. The second, denoted \textbf{ModDrop}, augments $\pi_{0.5}$ training with random modality token replacement, implementing the modality dropout strategy of~\cite{liu2017learning} on the $\pi_{0.5}$ backbone. Both baselines use the same architecture and training data as EGR.
\vspace{-10pt}

\paragraph{Metrics.} For Suite 1, we report $\Delta$, defined as the increase in per-frame action MSE under corrupted inputs relative to full inputs, evaluated only on task-relevant action dimensions $\mathcal{D}$: the robot base for NAV, and the torso and both arms for MAN. A formal definition is provided in the Appendix. For Suite 2, we report task success rate (SR) for both NAV and MAN segments. For MAN, we also report end-effector contact rate (EC), the fraction of trials in which the active arm contacts the focal object, and the minimum end-effector distance to the target (Dist). These auxiliary metrics capture partial progress even when SR is zero. Standard errors, confidence intervals, and significance tests for all reported numbers are provided in the Appendix.

\vspace{-5pt}
\subsubsection{Benchmark Construction}
\label{subsubsec:bench-construction}
\vspace{-5pt}

Our benchmark is built on BEHAVIOR-1K~\cite{li2023behavior}, using its official per-frame instance segmentation and skill annotations. We focus on the 12 tasks proposed by the second-place team in the BEHAVIOR-1K Challenge~\cite{bai2025openpicometcompetitionsolution}, of which 11 yield usable data (one excluded for truncated annotation). For each candidate skill segment, we apply three vision-evidence-based filters that align segments with the evaluation protocol of Sec.~\ref{subsubsec:evaluation}. NAV segments naturally rely on only the head camera; we keep NAV segments where the head is useful and at least one wrist is useless, evaluating them under Clean and SingleUseful. MAN segments are more complex and may a priori have any subset of cameras useful; for a uniform NoUseless / UsefulOnly evaluation across segments, we keep MAN segments where the head camera is useful and exactly one wrist is useless. After filtering, we re-decompose long-horizon demonstrations into segments and replay each demonstration to extract per-segment initial sim states. This yields 11 NAV + 36 MAN = 47 rollout-based segments. Filter statistics, threshold values, and replay procedure details are in the Appendix.

\vspace{-5pt}
\subsection{Diagnosing Modality Entanglement (Q1)}
\label{subsec:q1}
\vspace{-5pt}
\input{tables/table_suite1}

The vanilla $\pi_{0.5}$ rows of Table~\ref{tab:suite1}, in both the NAV and MAN panels, directly evidence both failure modes. On NAV, the SingleUseful condition corrupts both wrist cameras, which our filter has labeled as useless, while leaving the head camera (the only useful sensor) intact. Action deviation nonetheless increases by $+31.4\%$. Because NAV's SingleUseful condition simultaneously realizes NoUseless and UsefulOnly, it is a joint diagnosis of both failure modes. On MAN, corrupting the useless modality alone produces a $+12.7\%$ ratio, and keeping only the useful modality produces a $+158.5\%$ ratio. The first pair of numbers matches the nuisance sensitivity condition, and the third number matches the single-modality insufficiency condition. ModDrop shows comparable ratio patterns and higher absolute action deviation, indicating that random modality dropout does not eliminate this entanglement.

The rollout-based numbers in Table~\ref{tab:suite2} corroborate the diagnosis at the task-completion level. Vanilla $\pi_{0.5}$ on MAN drops from $12.5\%$ (Full) to $9.4\%$ (NoUseless) to $2.8\%$ (UsefulOnly); on NAV, from $42.3\%$ (Full) to $15.5\%$ (SingleUseful). Both suites converge: modality entanglement manifests at the action level (Table~\ref{tab:suite1}) and task-completion level (Table~\ref{tab:suite2}).

\vspace{-5pt}
\subsection{EGR Reduces Modality Entanglement (Q2)}
\label{subsec:q2}
\vspace{-5pt}
\input{tables/table_suite2}

The main result is the EGR row of Table~\ref{tab:suite2}. On MAN, EGR improves Full from $12.5$ to $16.4$ ($+31\%$ relative), NoUseless from $9.4$ to $16.5$ ($+75\%$), and UsefulOnly from $2.8$ to $6.1$ ($+120\%$). On NAV, Full is essentially preserved ($42.3 \to 40.9$) while SingleUseful rises from $15.5$ to $37.3$ ($+141\%$). Crucially, robustness gains do not come at the cost of clean-input performance, and MAN Full SR even improves substantially (12.5\% to 16.4\%, $+31\%$ relative), suggesting that evidence-gated regularization additionally helps the policy learn more task-relevant features even on clean inputs. The pattern of improvement matches the construction of the two losses, which target NoUseless and UsefulOnly directly. Table~\ref{tab:suite1} corroborates this: EGR's action-deviation ratios are consistently lower than vanilla and ModDrop (NAV SingleUseful: 31.4\%$\to$15.8\%; MAN UsefulOnly: 158.5\%$\to$121.8\%).

\vspace{-5pt}
\subsection{Real-Robot Validation (Q3)}
\label{subsec:q3}
\vspace{-5pt}

\begin{figure}[t]
  \centering
  \includegraphics[width=\linewidth]{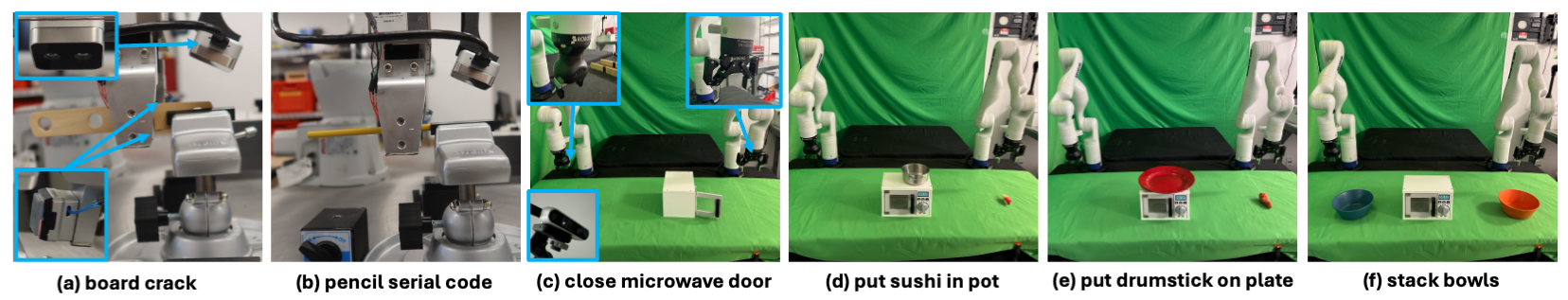}
  \caption{Real-robot platforms and tasks. \textbf{(a, b)} Vision-tactile platform. \textbf{(c-f)} Bi-manual platform. Sensor-detail zoom-ins are shown in (a) and (c).}
  \label{fig:hardware}
\end{figure}

We evaluate EGR on two real-robot platforms with substantially different embodiments and modality combinations (Fig.~\ref{fig:hardware}). The bi-manual platform consists of two Kinova arms and three RGB cameras, with the sensor layout shown in the inset of Fig.~\ref{fig:hardware}c. It is evaluated on three short-horizon tasks (Fig.~\ref{fig:hardware}c-e) and one long-horizon bowl-stacking task (Fig.~\ref{fig:hardware}f). The vision--tactile platform consists of a single MELFA ASSISTA arm, one RGB camera, and two GelSight tactile sensors, with the sensor layout shown in the inset of Fig.~\ref{fig:hardware}a. It is evaluated on two tactile feature-localization tasks (Fig.~\ref{fig:hardware}a-b), which emulate industrial inspection settings where visual information may be limited and tactile feedback provides task-relevant signal. Following the evaluation regimes in Sec.~\ref{subsubsec:evaluation}, we test the bi-manual platform, which has two useful sensors, under Clean, NoUseless, and UsefulOnly conditions. We test the tactile platform, for which the right GelSight is the sole useful sensor, under Clean and SingleUseful conditions. For both platforms, we also include a \textbf{RealDistractor} condition in which a physical out-of-distribution object is placed in the camera view, providing a non-synthetic probe of robustness.

\vspace{-5pt}
\subsubsection{Bi-Manual Vision Platform}
\label{subsubsec:bimanual}
\vspace{-5pt}

We run 4 tasks $\times$ 4 conditions $\times$ 10 trials per condition. Averaged across the four tasks, EGR improves Clean from 70\% to 80\% (+14\%), NoUseless from 52.5\% to 90\% (+71\%), UsefulOnly from 25\% to 72.5\% (+190\%), and RealDistractor from 30\% to 85\% (+183\%); see Fig.~\ref{fig:real_results}. The largest relative gain occurs on RealDistractor, where the distractor is a physical object the policy has never encountered, providing the strongest out-of-distribution signal among our conditions.

\vspace{-5pt}
\subsubsection{Vision-Tactile Platform}
\label{subsubsec:tactile}
\vspace{-5pt}

\begin{figure}[t]
  \centering
  \includegraphics[width=\linewidth]{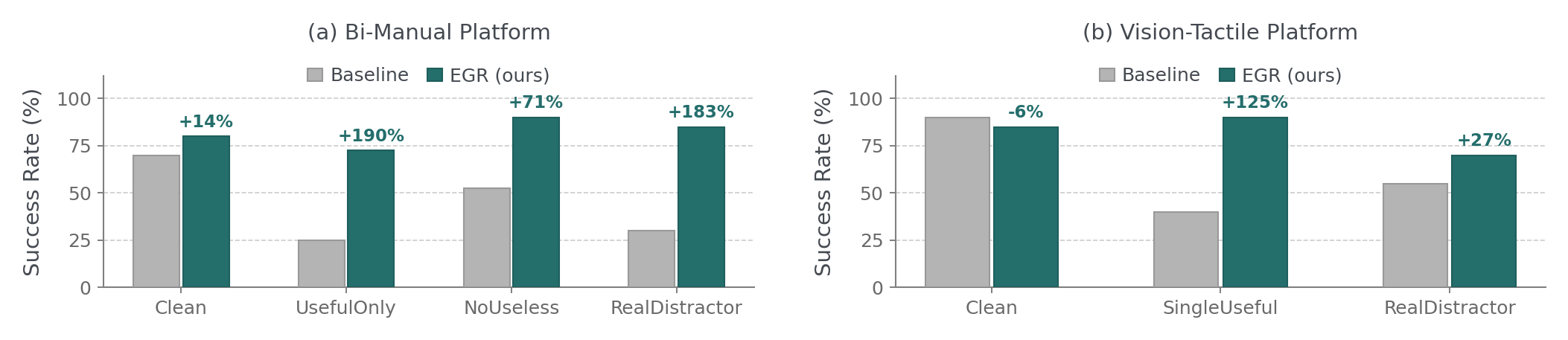}
  \caption{Real-robot results across two platforms (10 trials per condition). Each pair of bars compares the vanilla baseline (left, gray) and EGR (right, teal); relative gain is shown above each pair.}
  \label{fig:real_results}
\end{figure}

We run 2 tasks (pencil serial code, board crack) $\times$ 3 conditions $\times$ 10 trials per condition. Averaged across the two tasks, EGR preserves Clean within trial noise (90\% to 85\%, $-6\%$), improves SingleUseful from 40\% to 90\% ($+125\%$), and improves RealDistractor from 55\% to 70\% ($+27\%$); see Fig.~\ref{fig:real_results}. EGR improves performance on the tactile platform under both synthetic SingleUseful corruption and physical RealDistractor corruption, confirming that the framework's gains transfer to a fundamentally different sensor combination.

%% file: tables/table_suite1.tex
\begin{table}[t]
\centering
\setlength{\tabcolsep}{3pt}
\renewcommand{\arraystretch}{0.95}
\footnotesize
\caption{Suite 1 (inference-only): action deviation under per-modality perturbation. Ratio is the percentage increase relative to Full; lower is better. Bold marks the lowest ratio per column.}
\label{tab:suite1}
\begin{tabular}{lcccccccc}
\toprule
& \multicolumn{3}{c}{NAV} & \multicolumn{5}{c}{MAN} \\
\cmidrule(lr){2-4} \cmidrule(lr){5-9}
Method & Full & SingleUseful & Ratio & Full & NoUseless & Ratio & UsefulOnly & Ratio \\
\midrule
vanilla $\pi_{0.5}$ & 0.0616 & 0.0809 & $+31.4\%$ & 0.0020 & 0.0022 & $+12.7\%$ & 0.0052 & $+158.5\%$ \\
ModDrop & 0.0853 & 0.1172 & $+37.4\%$ & 0.0024 & 0.0026 & $+10.9\%$ & 0.0056 & $+137.7\%$ \\
\textbf{EGR (ours)} & 0.0780 & 0.0903 & $\mathbf{+15.8\%}$ & 0.0023 & 0.0024 & $\mathbf{+5.1\%}$ & 0.0051 & $\mathbf{+121.8\%}$ \\
\bottomrule
\end{tabular}
\end{table}

%% file: tables/table_suite2.tex
\begin{table}[t]
\centering
\setlength{\tabcolsep}{3pt}
\renewcommand{\arraystretch}{0.95}
\footnotesize
\caption{Suite 2 (rollout): task success rate (SR), end-effector contact rate (EC), and end-effector minimum distance to target (Dist). MAN columns report SR / EC / Dist. Bold marks the best per metric per regime.}
\label{tab:suite2}
\begin{tabular}{lcccccc}
\toprule
& \multicolumn{2}{c}{NAV ($n=11$)} & \multicolumn{3}{c}{MAN ($n=36$), SR$\uparrow$ / EC$\uparrow$ / Dist$\downarrow$} \\
\cmidrule(lr){2-3} \cmidrule(lr){4-6}
Method & Full & SingleUseful & Full & NoUseless & UsefulOnly \\
\midrule
vanilla $\pi_{0.5}$ & \textbf{42.27} & 15.45 & 12.50 / \textbf{52.78} / \textbf{0.25} & 9.44 / 47.22 / 0.26 & 2.78 / 22.78 / 0.45 \\
ModDrop & 40.00 & 15.45 & 7.50 / 40.97 / 0.29 & 6.81 / 38.89 / 0.29 & 2.78 / 22.36 / 0.47 \\
\textbf{EGR (ours)} & 40.91 & \textbf{37.27} & \textbf{16.39} / 50.00 / 0.26 & \textbf{16.53} / \textbf{53.89} / \textbf{0.25} & \textbf{6.11} / \textbf{30.14} / \textbf{0.37} \\
\bottomrule
\end{tabular}
\end{table}

%% file: sections/conclusion_and_limitation.tex
\vspace{-7pt}
\section{Limitations}
\label{sec:limitations}
\vspace{-7pt}

Our evidence instantiations derive from task structure: visible task-relevant geometry for vision and reference contact signatures for tactile. This covers settings in which the relevant signals can be specified a priori, but not cases where those signals must themselves be learned, such as in-hand object reorientation. Extending EGR to learn evidence from data, for example through information-theoretic objectives or self-supervised feature discovery, is a natural direction orthogonal to its regularization machinery. Our simulation benchmark also covers only 12 of the 50 BEHAVIOR-1K tasks so far; scaling to all 50 is straightforward future work using the same filtering and replay pipeline.

\vspace{-7pt}
\section{Conclusion}
\label{sec:conclusion}
\vspace{-7pt}
We presented Evidence-Gated Regularization (EGR), a training-time framework that mitigates modality entanglement in Vision-Language-Action policies. EGR uses a per-frame, per-sensor task-relevance signal to gate two state-conditional consistency objectives, requires no architectural change to the base policy, and adds zero inference cost. Across a BEHAVIOR-1K benchmark and two real-robot platforms with fundamentally different embodiments and sensors, EGR substantially improves success rates under both synthetic corruption and physical distractors.